\documentclass[10pt,twocolumn,letterpaper]{article}

\usepackage{cvpr}              

\usepackage{graphicx}
\usepackage{amsmath}
\usepackage{amssymb}
\usepackage{booktabs}

\usepackage[pagebackref,breaklinks,colorlinks]{hyperref}

\usepackage[capitalize]{cleveref}
\crefname{section}{Sec.}{Secs.}
\Crefname{section}{Section}{Sections}
\Crefname{table}{Table}{Tables}
\crefname{table}{Tab.}{Tabs.}

\def\cvprPaperID{68} 
\def\confName{EarthVision}
\def\confYear{2022}

\begin{document}

\title{Activation Regression for Continuous Domain Generalization with Applications to Crop Classification}

\author{Samar Khanna\\
Stanford University\\
{\tt\small sak296@cornell.edu}
\and
Bram Wallace\\
Salesforce \\
{\tt\small bw462@cornell.edu}
\and 
Kavita Bala\\
Cornell University \\
{\tt\small kb@cs.cornell.edu}
\and 
Bharath Hariharan\\
Cornell University \\
{\tt\small bharathh@cs.cornell.edu}
}
\maketitle

\begin{abstract}
Geographic variance in satellite imagery impacts the ability of machine learning models to generalise to new regions. In this paper, we model geographic generalisation in medium resolution Landsat-8 satellite imagery as a continuous domain adaptation problem, demonstrating how models generalise better with appropriate domain knowledge. We develop a dataset spatially distributed across the entire continental United States, providing macroscopic insight into the effects of geography on crop classification in multi-spectral and temporally distributed satellite imagery. Our method demonstrates improved generalisability from 1) passing geographically correlated climate variables along with the satellite data to a Transformer model and 2) regressing on the model features to reconstruct these domain variables. Combined, we provide a novel perspective on geographic generalisation in satellite imagery and a simple-yet-effective approach to leverage domain knowledge. Code is available at: \url{https://github.com/samar-khanna/cropmap}
\end{abstract}

\section{Introduction}
\label{sec:intro}

In a typical machine learning classification application, there are few guarantees of what knowledge is available at inference time. While it may be assumed that the data point on which one wishes to perform inference is within a predescribed distribution, a new data point is typically considered in isolation, with any additional descriptors primarily being categorical in nature (such as discrete domains in the problem of domain generalization).


To profit from increased information during inference, one school of thought advocates calculating gradients on the provided information. Such approaches include test-time training (typically done in a self-supervised manner), semi-supervised learning (when the target domain/distribution is available), or transductive learning (in the case of an even stronger assumption that the data we want to perform inference on itself is available during training time \cite{testtimetraining,s4l,pseudolabeling,joachimstransductive}. These methods, while effective in certain scenarios, can suffer from prohibitive computational cost.


A desired technique might instead be to leverage the extra information associated with a datapoint at inference time, in a \textit{generalization} setting, where only a single forward pass of the network is performed. For example, Multisource Domain Adversarial Networks (MDANs) train a network to yield a feature space where domain membership is \textit{unrecoverable} \cite{mdan,dann}. In essence, the goal is to achieve features which are agnostic to all information except their defined class membership. This is logical when domains are few and categorically defined as in the studied fields of domain adaptation/generalization, but in the limit of increasingly numerous incrementally different domains both the intuition and mathematical machinery fall apart.


Consider the problem of \textit{geographic} generalization, where some classification task is desired to perform in any given location.
While the differences between locations may typically be trivial at small scales, the generalization gap grows to be large in a primarily continuous manner.
Categorizing the globe into arbitrary bins as domains would not only break this reality, but also rely on a combination of hand-crafted reasoning and/or hyperparameter selection.
Instead, a notion of \textit{continuous} domains can be employed; the domain changes encountered will be of a densely populated range and at inference time the domains will be an extrapolation of those included in the training data.
In such a setting, the question arises whether domain-masking favored by approaches like MDAN is indeed the right approach. Within a continuously varying domain, knowledge of a point's domain may instead provide valuable knowledge and allow a model to adapt execution based on the recognition of familiar or unfamiliar settings. We consider this problem in the context of land-use crop classification.

Agriculture is performed through the world\cite{almanac2020}.
What crops one might observe in a given part of the world can depend on a number of factors: atmospheric, geologic, economic, etc\cite{hardiness_zones}. 
At an extremely basic level, the United States Department of Agriculture (USDA) hardiness zones delineate regions by the minimum temperatures throughout the year, helping to advise which plants can survive in certain regions\cite{hardiness_zones}.
For example, succulents such as aloe cannot survive in cold regions, while many leafy greens such as cabbage can endure freezing temperatures.
Other factors (such as soil, elevation) matter as well, but climate is one of the chief determinants to know what might be planted \cite{hu2003climate}.

Domain generalization is a logical way to frame this setting, given how land-use and crop labels are typically collected and organized.
Many highly developed areas of the world conduct large-scale publicly available surveys catalogging large-scale land-use.
Examples of such surveys include the USDA Cropland Data Layer or Canada AAFC Crop Inventory\cite{usda_cdl,canada_crops}.
Such censuses yield dense (30m resolution) annotations with fine-grain crop information across the entire country.
In these cases, in-domain classification itself is unnecessary as the labels have already been collected.
So by its nature, for crop classification to be useful there typically must be some sort of domain shift. 
This could result from:
\begin{itemize}
\item Year-to-year generalization: enabling early prediction on the subsequent years crops. This scenario is of limited use depending on the timing of the annual census.
\item Interpolative generalization: if the labels obtained are sparse, then filling in the regions between them might be of interest. This is an unusual case that does not present itself in typical datasets.
\item Geographical extrapolation: this is the scenario that we consider of the highest interest. Given labels for one geographic region but not another, domain generalization can be performed. This comes with an inherent domain shift, which can partially be characterized by climate data.
\end{itemize}

Land-use classification is typically done via aerial (satellite) imagery.
This presents interesting avenues of exploration, as there is a massive abundance of available satellite and geospatial data which can easily be reconciled and aligned with a given set of aerial images.
This is a key deviation from the typical generalization settings.
Topographical, climatic, air quality, and hydrology maps are just a sparse sampling of the available geospatial data available; this is in addition to the wide variety of imaging devices such as LandSat, Sentinel, MODIS, and NAIP satellites\cite{landsat,sentinel,modis,naip}.
For a given geographical point, there is a vast amount of potentially relevant data available, much more than could be expected in a typical benchmark setting.
In this work, we particularly focus on the impact of climate information.

A solution to the problem of geographically-generalizing in-season crop classification would greatly enhance global food security as crop yield predictions in food-insecure countries could highlight potential food shortages months before they would otherwise be realized.
This problem presents a different type of domain generalization, which we term \textit{continuous} domain generalization (CDG).
CDG presents intriguing technical aspects, particularly how to leverage associated information efficiently at inference time.
In this setting, we present a method to incorporate climate information to improve geographic generalization across continously varying domains using a technique we dub \textbf{A}ctivation \textbf{R}egression for \textbf{C}ontinuous \textbf{Do}main \textbf{G}eneralization, or \textbf{ARCDoG}.

\section{Related Work \& Background}

\subsection{Domain Generalization}

Domain generalization (DG) is the problem of training neural networks that generalize to domains that contain the same content as the training data with different statistical properties \cite{dg_survey}.
Examples of domain generalization in computer vision include PACS (generalization between photos, art, cartoons, and sketches), Office-Home (between art, clipart, product, and real-world), and Digits (between handwriting and signs)\cite{dg_survey}.
Methods broadly fall into two categories: those that train individual domain-specific network modules and those where all parameters are shared.
The latter category includes approaches such as Episodic Training for Domain Generalization and MetaReg, where episodic/meta-learning approaches are employed to train each domain's subnetwork to operate on each other domains\cite{epidg,metareg}.
This family of methods is poorly defined for \textit{Continuous} DG, as it encounters the previously noted problems associated with arbitrary categorical divisions.

Of more applicability is the set of methods that learn a single unified representation across all domains; these offer greater promise to our problem setting.
One such subclass of techniques employs self-supervised losses on the target domain to regularize the feature space.
An example of such a method is JiGen, which employs the Jigsaw self-supervised loss to learn how to generalize in a self-supervised manner\cite{jigen}.
This subclass again is less aplicable to aerial imagery and self-supervision is under studied, with initial forays such as Tile2Vec being much less applicable to the data we are working with as described in \cref{sec:method}\cite{tile2vec}.
Additionally, such methods by design target specific domains to generalize to, instead of providing true generalization (this is true of most semi-supervised settings, with transductive learning carrying these assumptions to the extreme of training on the unlabeled datapoints we will ultimately attempt to classify).

A group of methods that offers much promise, however, is that of adversarial adaptation, the primary works being Domain Adversarial Neural Networks and Multi-Domain Adversarial Networks (DANN and MDAN respectively), with MDAN being the multi-domain extension of DANN\cite{dann,mdan}.
In MDAN, a classifier is trained across all source domains to perform classification while yielding a feature space where the domain of a datapoint is unrecoverable.
A simplified version of MDAN is as follows, consider a feature extractor $\theta$, classification heads (linear or multi-layer perceptrons, MLPs) $\phi_{class}$ and $\phi_{domain}$, and data $x_i^{(k)}$ being the $i^{th}$ datapoint of the $k^{th}$ domain with corresponding labels $y_i^{(k)}$ drawn from datasets $X$ and $Y$.  Let $L_{cls}$ denote a classification loss, typically cross-entropy and $\vec{e}^{(k)}$ the $k^{th}$ standard basis vector. 
The (simplified) training objective of MDAN is:
\begin{align}
\begin{split}
\min_{\theta, \phi_{class}} \bigg[ \mathop{\mathbb{E}}_{X,Y} \bigg[ & L_{cls}(\phi_{class}(\theta(x_i^{(k)})), y_i^{(k)}) - \\
& c \cdot \min_{\phi_{domain}} \left[  L_{cls}(\phi_{domain}(\theta(x_i^{(k)})), \vec{e}^{(k)}) \right] \bigg]\bigg]
\label{eq:mdan}
\end{split}
\end{align}
\noindent
where $c$ is a weighting hyperparameter. Here the objective is to maintain good classification accuracy while competing against an adversarially trained network attempting to discriminate the individual source domains from their associated features.
While the categorical delineation of domains does not lend itself to CDG, we present ARCDoG in \cref{sec:method} which adapts MDAN to the continuous setting using a closed-form linear regression in lieu of the adversarial network $\phi_{domain}$.

\subsection{Machine Learning on Satellite Data}

Satellite imagery has recently seen increasing interest from the machine learning community, with benchmarks such as BigEarthNet, DeepGlobe, and UC Merced Landuse being introduced to measure progress on classification tasks in a similar manner to that of research on the ImageNet benchmark or CIFAR-10/100\cite{bigearthnet,deepglobe,merced,imagenet,cifar}.
While such benchmarks have encouraged rapid iterative development in the broader field of computer vision, the problem of benchmark versus in-the-wild application performance are accentuated for aerial imagery.
One aspect of this problem is curation: image datasets are typically highly curated with images being drawn from a very narrow distribution and having limited content; such models are then often brittle in practice.
In the application of satellite imagery, typically inference will be performed across a large contiguous swath of land, in which a model must handle \textit{all} data, not just a relevant curated subset.
Another aspect of the difficulty in a satellite imagery benchmark is the \textit{availability} of satellite imagery.
The whole globe is visible by satellites, meaning that problems must be formulated carefully to maintain relevance across training-validation-testing splits (see our specific discussion regarding crop classification in \cref{sec:intro}).

Satellite ``imagery" is often in fact limited in spatial resolution. Publically available instruments with high temporal frequency typically sample pixels on the order of 30 meters by 30 meters.
Generally, there is a tradeoff between temporal and spatial resolution, for example the National Agriculture Imagery Program (NAIP) acquires data with 1 meter resolution, but on a 3-year cycle\cite{naip}.
For large-scale classification problems, temporal resolution becomes of utmost importance \cite{tempImportant}, especially since stochastic cloud cover can render any given satellite capture unusable. 
Moreover, transformer architectures have shown their efficacy on temporal satellite data \cite{cropTransformer}. 
Therefore, lower spatial resolution data is operated on in a collated timeseries, rather than a static image as is often considered in computer vision.
For classification tasks where the property of interest can have pixel-scale variations, this sequence is of a single pixel location, forming a multivariate timeseries. 

\subsection{Satellite Data}
\label{subsec:sat_data}


Part of the recent interest in remote sensing stems from the abundance of geospatial data available in addition to standard imagery, such as ozone measurements, night light intensities, elevation maps, daily weather history, and more\cite{earth_engine_datasets}.
We employ three particular data sources in this work.

\paragraph{USGS Landsat 8 Surface Reflectance Tier 1}

\begin{figure}
  \centering
   \includegraphics[height=250pt]{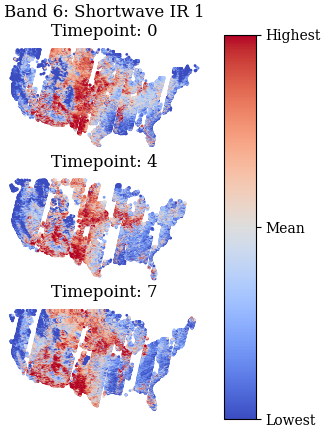}
   \caption{Landsat 8 band B6 (Shortwave Infrared 1) visualised across the US across three different timepoints. Here timepoint 0 is the first snapshot in April, timepoint 4 is the first snapshot in June, and timepoint 7 is the last snapshot in July. }
   \label{fig:landsat_near_ir}
\end{figure}

\begin{figure}
    \centering
    \includegraphics[width=\linewidth]{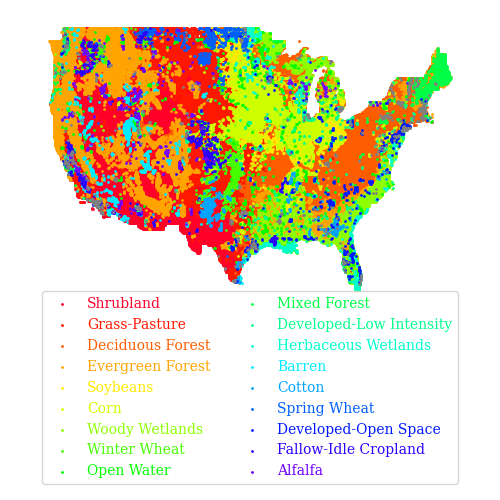}
  \caption{18 most common Cropland Data Layer (CDL) classes.}
  \label{fig:cdl}
\end{figure}
\begin{figure}
  \centering
  \includegraphics[width=0.95\linewidth]{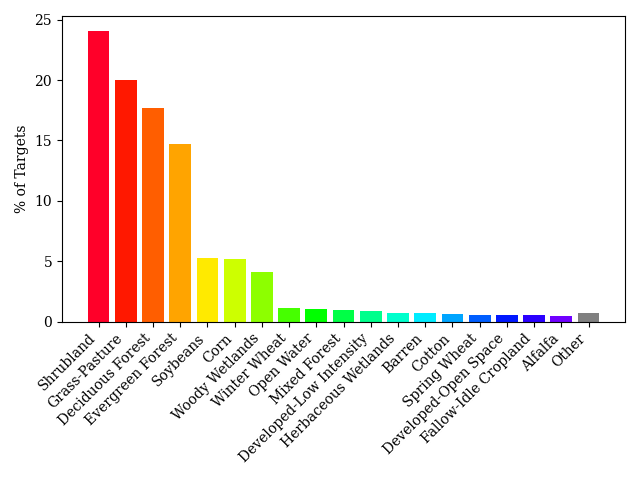}
  \caption{Histogram of crop class counts across the US}
  \label{fig:class_counts}
\end{figure}

The Landsat mission has been conducted by the USDA and NASA for almost 50 years with the goal of acquiring large quantities of detailed satellite images for varied scientific use\cite{landsat}.
Landsat 8 is a satellite launched in February 2013 which collects 9 hyperspectral bands of surface reflectance, detailed in \cref{tab:landsat}.
The temporal resolution of Landsat 8 is relatively high, it captures the entire globe of imagery every 16 days, this comes with a spatial resolution tradeoff, however.
The \textit{spatial} resolution of Landsat is $\frac{900m^2}{pixel}$, or $30m \times 30m$. Given the lack of discernible visual features, recognizing crops and classifying land-use is therefore a challenging non-textural problem.

\begin{table}
  \centering
  \begin{tabular}{c c c}
    \toprule
    Band ID & Wavelength & Band Name  \\
    \midrule
    B1 & 0.435-0.451$\mu m$ & Ultra Blue \\
    B2 & 0.452-0.512$\mu m$ &  Blue \\
    B3 & 0.533-0.590$\mu m$ & Green \\
    B4 & 0.636-0.673$\mu m$ & Red \\
    B5 & 0.851-0.879$\mu m$ & Near Infraread \\
    B6 & 1.566-1.651$\mu m$ & Shortwave Infrared 1 \\
    B7 & 2.107-2.294$\mu m$ & Shortwave Infrared 2 \\
    B10 & 10.60-11.19$\mu m$ & Temperature \\
    B11 & 11.50-12.51$\mu m$ & Temperature \\
    \bottomrule
  \end{tabular}
  \caption{Spectral bands of LandSat-8}
  \label{tab:landsat}
\end{table}

The reason we use this data, however, is the aforementioned problem of cloud cover. There are very few high-temporal-frequency data sources available, and high frequency is critical to prevent pixels from being obscured by clouds in all captures.
As we are trying to perform crop classification \textit{within the growing season}, high temporal frequency is even more important.
In this work, we sample imagery bi-monthly from April to August.
This window captures the entirety of the growing season in much of the US while still allowing for pre-harvest prediction.
An example timeseries of images for the near-infrared channel (B5) is shown in \cref{fig:landsat_near_ir}, note the variations across time, missing patches caused by cloud cover, and diagonal striping from separate Landsat collection passes.

\paragraph{USDA Cropland Data Layers (CDL)}

Since 2008, the USDA has created a crop-specific landuse map of the entire continental US annually at a resolution of $900m^2 /pixel$ (same as Landsat)\cite{usda_cdl}.
These annotations heavily rely on on-the-ground censusing, as well as human labeling of satellite images.
There are over 100 classes included, with the large majority being fine-grained species of crops.
A visualization is shown in \cref{fig:cdl}.

\paragraph{WorldClim BIO Variables V1}

\begin{figure}
  \centering
   \includegraphics[width=\linewidth]{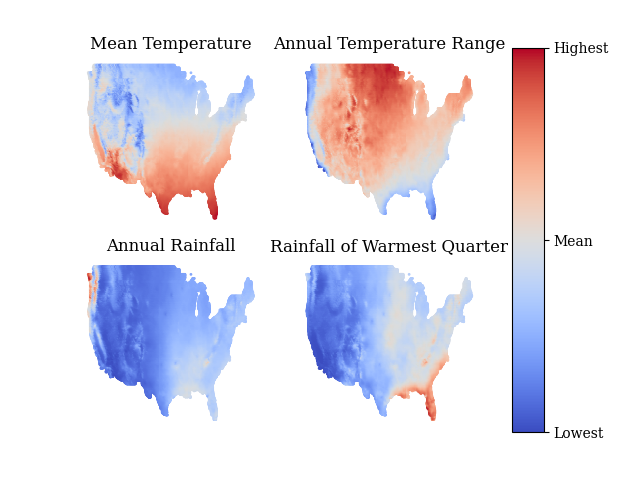}
   \caption{Some of the WorldClim BIO variables visualised.}
   \label{fig:climate}
\end{figure}

This dataset is a collection of bioclimatic variables averaged over 30 years\cite{worldclim}.
Spatial resolution is approximately $900m \times 900m$, while this is much lower than the other data layers incorporated, the relatively smooth nature of climate variations means that WorldClim can be upsampled with little consequence.
There are 19 different quantities reported, consisting of a variety of temperature/precipitation averages and spreads as well as relations between the two.
See \cite{worldclim} for a complete description of bands and \cref{fig:climate} for several visualized bands across the continental USA.

\subsection{Closed-form Linear Regression in Machine Learning}
\label{sec:pseudoinverse}

Our work uses backpropagation through closed-form least-squares linear regression via the pseudoinverse\cite{pseudoinverse}.
The pseudoinverse of a full-rank matrix $A$ is 
\begin{equation}
A^{+}=(A^{*}A)^{-1}A^{*}
\end{equation}
Assuming invertibility (which while a strong assumption is generally true for neural network activations albeit with poor conditioning), this construction is numerically stable as the inversion is of a Hermitian matrix.
Given a linear system $Ax=b$, $x=A^{+}b$ is an optimal solution with residual $b - AA^{+}b$.
There are other instances of closed-form linear regression being employed in deep learning literature, such as Feature Reconstruction Networks\cite{frn}.
The Maximal Coding Rate Reduction loss can be used analogously in a classification setting\cite{mcr2}.

\section{Method}
\label{sec:method}
We describe our data generation process in \ref{subsec:data}, the optimization objective in \ref{subsec:arcdog}, and training/testing details in \ref{subsec:details}.

\subsection{Data}
\label{subsec:data}

\begin{figure*}
  \centering
  \includegraphics[width=\linewidth]{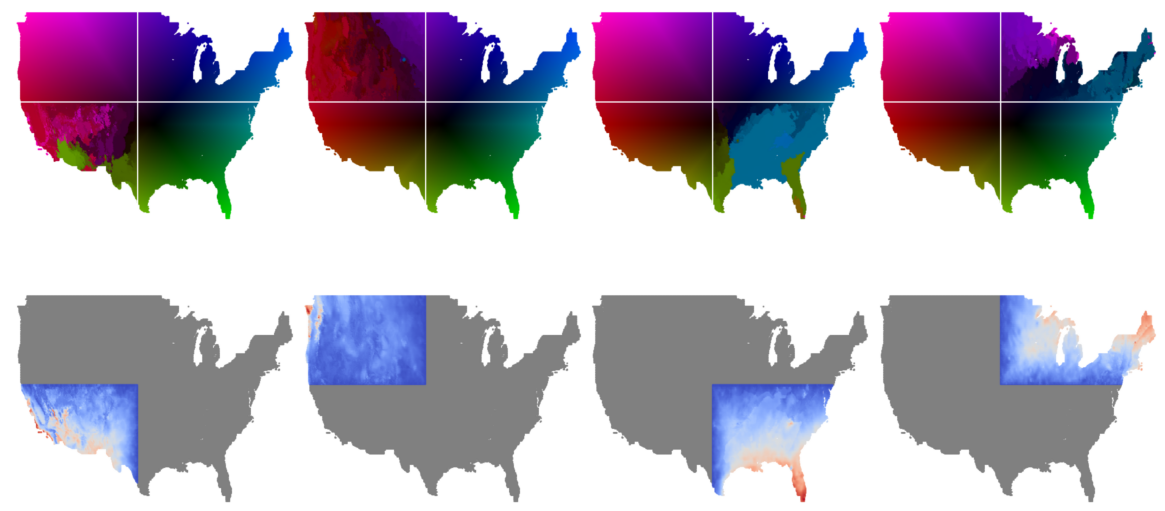}
  \caption{1-nearest neighbour climate features for each of the 4 regions. The top row colours each point in the testing domain region with the colour of the most similar feature vector from the other 3 training regions. The bottom row plots a heat-map of the raw distance to the nearest feature from the 3 training regions (red is further, blue is closer). The features are the normalised climate variables.}
  \label{fig:climate_knn}
\end{figure*}

We sample the continental US across a grid of spatial resolution 1 mile, this results in a sparse grid of approximately 3 million coordinates.
At each point, Landsat 8 imagery, Bioclim Variables, and CDL are sampled (the latter two once, the former over the span of April to August resulting in 8 timepoints of 9 channels each).
We partition this grid across the median latitude and longitude into 4 quadrants, as shown in \cref{fig:climate_knn}.


\begin{figure*}
    \centering
    \begin{subfigure}{0.25\linewidth}
        \includegraphics[width=\linewidth]{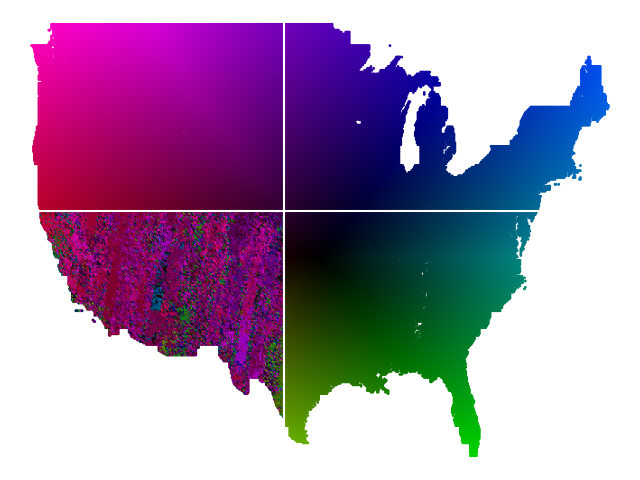}
        \caption{Region 0 (South West)}
        \label{fig:knn0}
    \end{subfigure}%
    \begin{subfigure}{0.25\linewidth}
        \includegraphics[width=\linewidth]{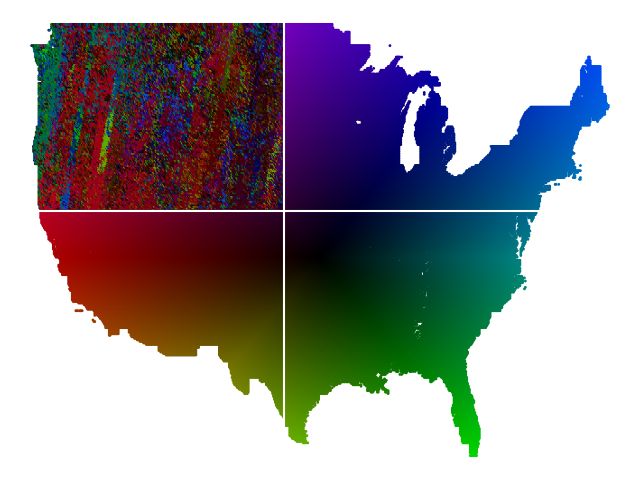}
        \caption{Region 1 (North West)}
        \label{fig:knn1}
    \end{subfigure}%
    \begin{subfigure}{0.25\linewidth}
        \includegraphics[width=\linewidth]{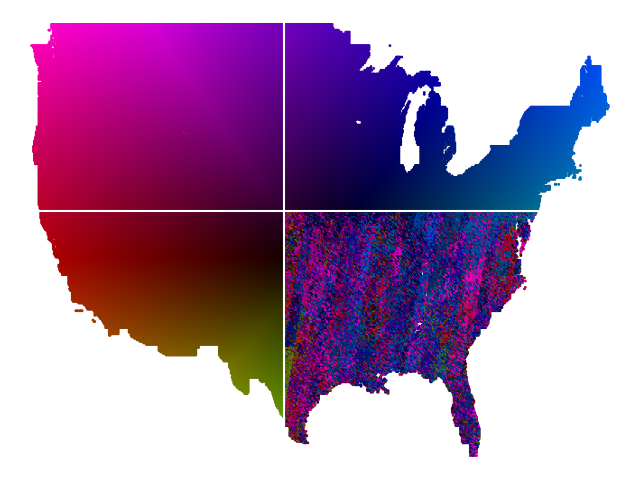}
        \caption{Region 2 (South East)}
        \label{fig:knn2}
    \end{subfigure}%
        \begin{subfigure}{0.25\linewidth}
        \includegraphics[width=\linewidth]{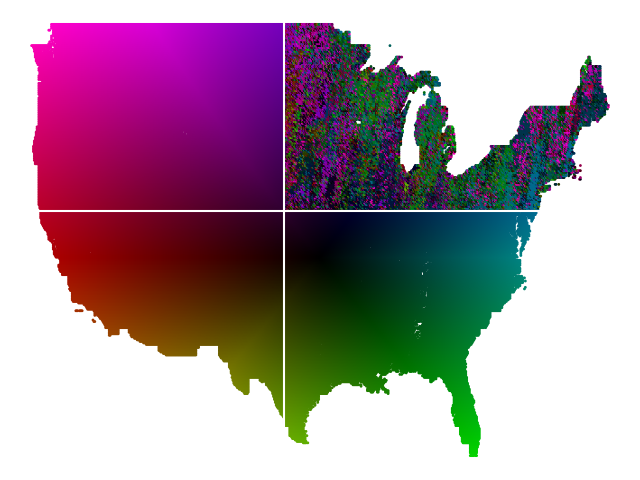}
        \caption{Region 3 (North East)}
        \label{fig:knn3}
    \end{subfigure}
    \begin{subfigure}{0.25\linewidth}
        \includegraphics[width=\linewidth]{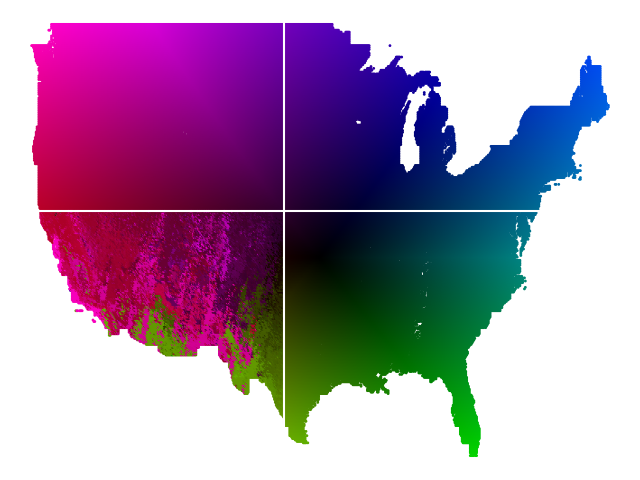}
        \caption{Region 0 (South West)}
        \label{fig:knn0clim}
    \end{subfigure}%
    \begin{subfigure}{0.25\linewidth}
        \includegraphics[width=\linewidth]{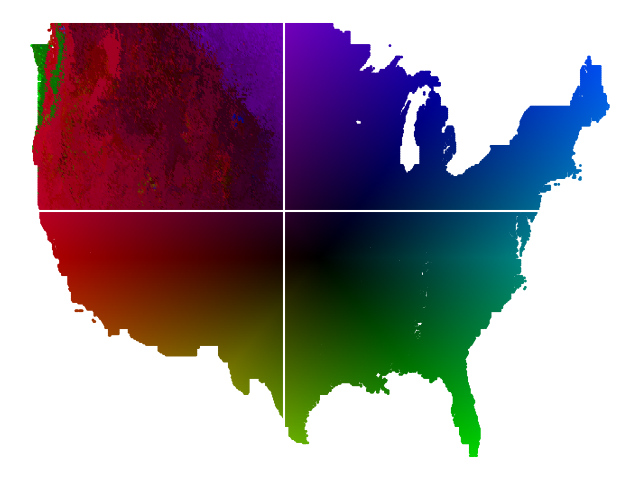}
        \caption{Region 1 (North West)}
        \label{fig:knn1clim}
    \end{subfigure}%
    \begin{subfigure}{0.25\linewidth}
        \includegraphics[width=\linewidth]{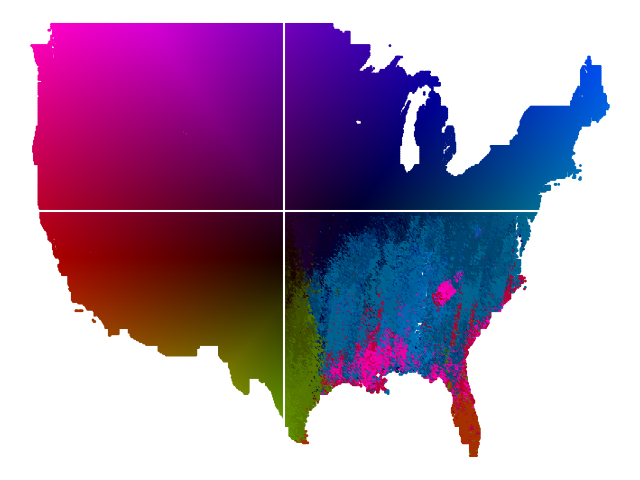}
        \caption{Region 2 (South East)}
        \label{fig:knn2clim}
    \end{subfigure}%
    \begin{subfigure}{0.25\linewidth}
        \includegraphics[width=\linewidth]{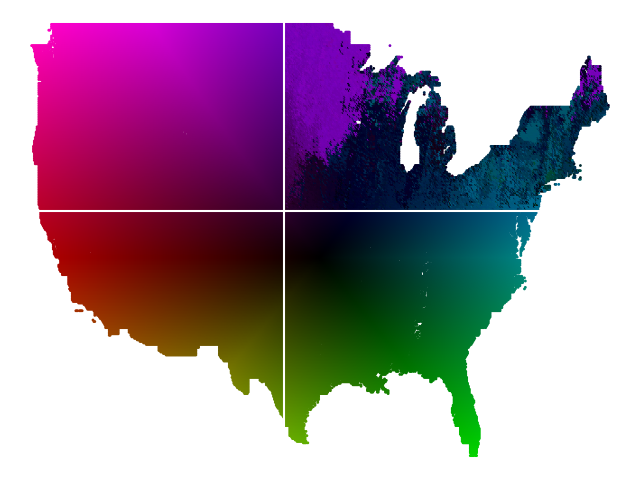}
        \caption{Region 3 (North East)}
        \label{fig:knn3clim}
    \end{subfigure}
    \caption{1-nearest neighbour features for each of the 4 regions. In each figure, we colour each point in the testing domain region with the colour of the most similar feature vector from the other 3 training regions. The top row uses  feature vectors are from the final layer of the baseline Transformer model without climate variables as input, and the bottom row uses feature vectors from the Transformer model with climate variables as input. Euclidean distance was used for kNN. For the benefit of visualisation, all grid points of the training quadrants were input to the transformer, not just those corresponding to the designated crop classes. See \ref{subsec:data} for details.}
    \label{fig:knn}
\end{figure*}

\begin{figure*}
    \centering
    \includegraphics[width=\linewidth,height=175pt]{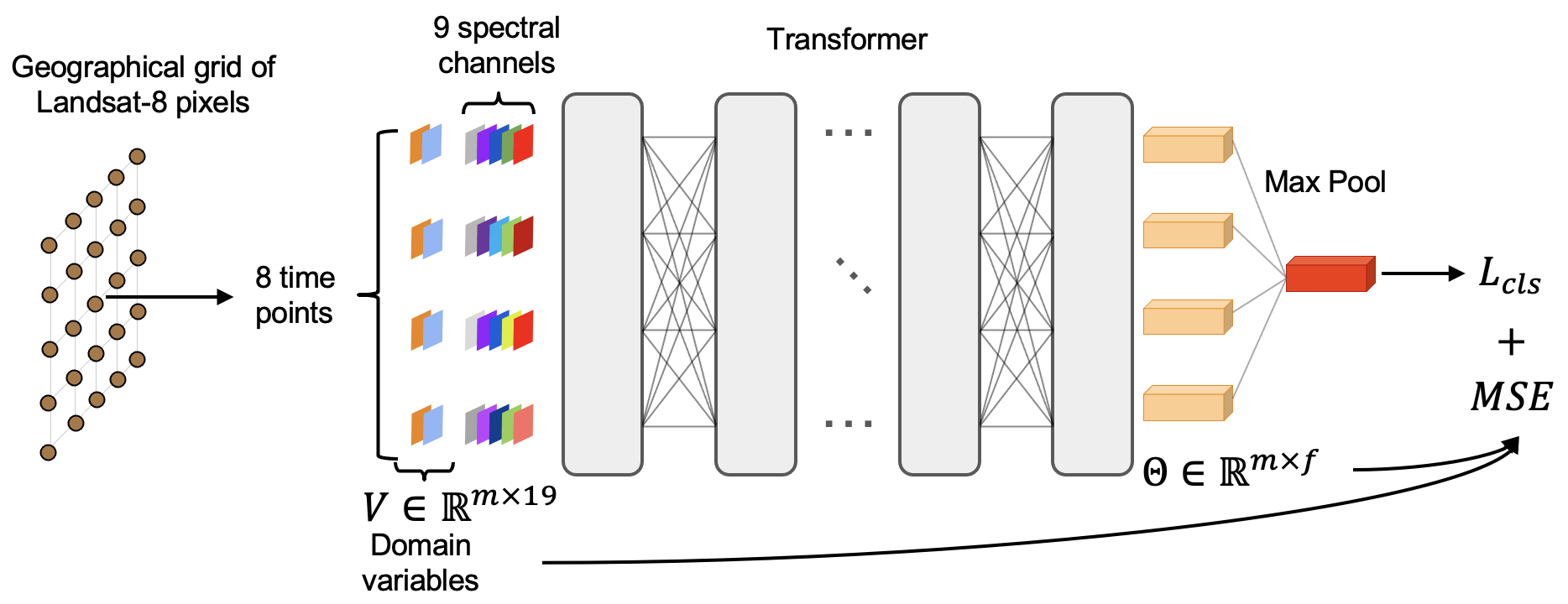}
    \caption{Method described in section \cref{sec:method}. The grid of pixels is sampled at roughly 1609m $\sim$ 1 pixel across the lower 48 states of the continental US. $V$ is a climate descriptor for each location on the grid, copied over for each of the 8 time points across April-July (2 per month). We consider 9 spectral channels of Landsat, see \cref{tab:landsat}. The loss is a combination of the typical classification cross entropy loss and MSE of closed-form linear regression of the climate descriptor from the final representation \cref{eq:arcdog}.}
    \label{fig:transformer}
\end{figure*}

These quadrants serve to partition the domains we use to measure our method's domain generalization capability.
Each quadrant can serve as a target domain, with the source domain being the other three sections of the map.
This 3-to-1 generalization allows for four different evaluations of a method in leave-one-out style.
Note how the generalization gap is indeed continuous as previously described, we cement this notion in \cref{fig:climate_knn} where a heatmap shows the nearest neighbor matches to all points in each generalization quadrant.
While the climate is locally constant while crossing the boundary of a generalization region, it diverges increasingly as the distance grows with unpredictable patterns emerging.
As seen in \cref{fig:class_counts}, most of the class labels are non-crop classes (eg: shrubland, forest, etc.) and are distributed across wide portions of the US \cref{fig:cdl}. These classes are weakly correlated with geographic location, and therefore might serve to confuse the model. Moreover, since our goal is to predict crop classes, we modify our dataset to only include datapoints belonging to a set of 25 curated crop classes, discarding all other data points.
Additionally, we see in \cref{fig:class_counts} that the crop distribution varies widely between generalization regions as well.
This problem of continuous domain generalization presents an intriguing setup, which we demonstrate a tool for in the following section.

\subsection{ARCDoG: Activation Regression for Continuous Domain Generalization}
\label{subsec:arcdog}

Recall the MDAN objective in \cref{eq:mdan} where the goal  is to optimize classification performance while simultaneously obfuscating the source domains of representations.
Note that if $c \le 0$, then the auxiliary objective would \textit{encourage} classifiability of the source domains.
In our setting, the domain is characterized not by a categorical definition with one-hot associations, but by the climate variables at a location which form a vector which we denote as $\vec{z_i}\in\mathbb{R}^{19}$.
Since this quantity is a real-valued vector instead of a binary classification target, a regression instead of classification loss is appropriate.

Adversarial training is notoriously unstable\cite{gan_survey}.
As such, we choose to calculate the "adversarial`` loss, in a closed-form parameter-free manner.
Specifically, consider the loss term
\begin{equation}
 \min_{\phi_{domain}}  \left[ \mathop{\mathbb{E}}_{X,Y}  \left[  L_{cls}(\phi_{domain}(\theta(x_i^{(k)})), \vec{e}^{(k)}) \right] \right]
\label{eq:test}
\end{equation}
which measures the separability of representations by source domain.
To measure this quantity, a head MLP $\phi_{domain}$ must be trained with a competing goal to that of $\phi_{class}$ and thus requiring tradeoffs in the learning of $\theta$.
If the domains were parametrized continuously instead of categorically, by a vector $\vec{v}^{(k)}$, then this loss could be adapted to use mean-squared error (MSE) loss as:
\begin{equation}
 \min_{\phi_{domain}}  \left[ \mathop{\mathbb{E}}_{X,Y} \left[ MSE(\phi_{domain}(\theta(x_i^{(k)})), v^{(k)}) \right] \right]
\end{equation}
As we do not delineate source domains, we shall discard the domain indexing $k$ going forward and let $\vec{v_i}$ describe the descriptor of the domain for the $i^{th}$ datapoint.

When $\phi_{class}$ is an MLP or similarly analytically complex neural network, then it must be learned through gradient descent in conjunction with the other networks to approximate $\min_{\phi_{domain}}$
If $\phi_{domain}$ is a \textit{linear} regression model, however, then the calculation of the minimum is much simpler as it exists in closed-form solution across a single batch.
Specifically, given a batch of indices of size $m$, $\{i_j\}_{j=1}^{m}$ let $\Theta = \ \left( \theta(x_{i_j})^T \right)_{j=1}^{m} \in  \mathbb{R}^{m \times f} $ denote the features of the batch  and the climates as $ V= \left( v_{i_j}^T \right)_{j=1}^{m} \in  \mathbb{R}^{m \times 19}$ where $f$ is the feature dimensionality output by $\theta$.
Then letting $\Phi \in \mathbb{R}^{f \times 19}$ be the linear regressor matrix and using the pseudoinverse as described in \cref{sec:pseudoinverse}
\begin{align}
 \min_{\phi_{domain}} & \left[ MSE(\phi_{domain}(\Theta), V) \right] = \\
&\min_{\Phi}  \left[ MSE(\Theta \Phi, V) \right] =\\
& MSE(\Theta \Theta^{+} V, V) 
\end{align}
enabling closed-form calculation of the second term of the loss.

Reincluding the classification loss term and accounting for our continuous setting, we arrive at the loss we employ for ARCDoG:
\begin{equation}
\begin{split}
L_{ARCDoG} = \min_{\theta, \phi_{class}} \bigg[ \mathop{\mathbb{E}}_{X,Y} \bigg[  &L_{cls}(\phi_{class}(\theta(x_i^{(k)})), y_i^{(k)}) - \\
&c \cdot MSE(\Theta \Theta^{+} V, V) \bigg] \bigg]
\label{eq:arcdog}
\end{split}
\end{equation}
Observe that $c > 0$ encourages the climate $V$ to be \textit{un}recoverable from the features $\Theta$ while $c < 0$ guides the climate to be recoverable.
This ARCDoG loss ports the intuition of MDAN into the continuous domain realm, while doing away with the difficulties of adversarial training by employing a simple linear $\phi_{domain}$.

\subsection{Details}
\label{subsec:details}
 
The landsat data input into our model is a timeseries of 8 points with 9 channels each resulting in $x_i \in \mathbb{R}^{8 \times 9}$, in some experiments we incorporate the climate as input as well, appending $\vec{v}_i  \in \mathbb{R}^{19}$ to each timepoint resulting in $x'_i \in \mathbb{R}^{8 \times 28}$. 

For $\theta$, a transformer architecture inspired by \cite{cropTransformer} is employed on single pixel vectors \ref{fig:transformer}, $\phi_{class}$ is a linear head\cite{transformers}.
Models are trained on three source domains at a time with 10\% of the training data being used for loss validation.
An Adam optimizer with initial learning rate of $1e-3$ and learning rate reduction on plateau schedule is used with a factor of $0.1$ and patience of 5 epochs (the criterion being minimization of validation loss).
Upon the 3rd reduction, training is terminated and early stopping is employed (the best version of the model with respect to validation loss is reloaded and used).
Cross entropy loss is used for classification.
The MSE loss of the second term is normalized by the norm of $V$, resulting in a loss term of
\begin{equation}
\frac{||V-\Theta \Theta^{+} V||_2}{||V||_2}
\end{equation}
this modification helps the two loss terms of $L_{ARCDoG}$ to be on the same scale.
A batch size of 4096 is used.

We note that while training of our model is more expensive, compared to a baseline of a vanilla supervised transformer there are no extra parameters and at inference time computational cost is unaffected. 

\section{Results}
\label{sec:results}

Our main experiment trains the transformer model on 3 of the input quadrants and tests it on the held-out 4th quadrant. The results are in table \ref{tab:results}. We also investigate the impact of the different types of climate variables in the model's performance. The WorldClim BIO variables are roughly divided among those related to temperature (\texttt{bio01}-\texttt{bio11}) and precipitation (\texttt{bio12}-\texttt{bio19}). We ablate the temperature and precipitation variables, separately, as inputs to the model, and these results are recorded in table \ref{tab:ablate}.

\begin{table}[t]
  \centering
  \begin{tabular}{c c c c c}
    \toprule
    Method & 0 & 1 & 2 & 3  \\
    \midrule
    Baseline & 0.386 & 0.335 & 0.484 & \textbf{0.526} \\
    $c = -1$ & 0.402 & 0.382 & 0.572 & 0.503 \\
    $c = -0.1$ & 0.406 & 0.381 & 0.560 & 0.490 \\
    $c = 0$ & \textbf{0.421} & 0.386 & 0.571 & 0.480 \\
    $c = 0.001$ & 0.413 & \textbf{0.394} & 0.578 & 0.475\\
    $c = 0.01$ & 0.420 & 0.391 & \textbf{0.600} & 0.468 \\
    $c = 0.1$ & 0.412 & 0.391 & 0.575 & 0.473 \\
    $c = 1$ & 0.409 & 0.386 & 0.555 & 0.477 \\
    \bottomrule
  \end{tabular}
  \caption{Mean accuracy results. Each column denotes the test region IDs, after the Transformer model was trained on all other regions. In each row (except the baseline), the model was inputted the 9 spectral channels as well as the 19 climate variables. The baseline was given only the 9 spectral channels. The value $c$ determines the strength of the correlation enforced in the $L_{ARCDoG}$ loss \ref{eq:arcdog}. Each number is the result of averaging over 5 trials. }
  \label{tab:results}
\end{table}

\begin{table}[t]
  \centering
  \begin{tabular}{c c c c c}
    \toprule
    Method & 0 & 1 & 2 & 3  \\
    \midrule
    Baseline & 0.386 & 0.335 & 0.484 & 0.526 \\
    All climate input & 0.421 & \textbf{0.386} & 0.571 & 0.480 \\
    Temperature only & \textbf{0.429} & 0.347 & \textbf{0.574} & 0.526 \\
    Precipitation only & 0.368 & 0.360 & 0.500 & \textbf{0.556} \\
    \bottomrule
  \end{tabular}
  \caption{Mean accuracy results. Each column denotes the test region IDs, after the Transformer model was trained on all other regions. Each result is with a regression weight of $c=0$. Each number in the table is the result of averaging over 5 trials. }
  \label{tab:ablate}
\end{table}

\section{Discussion}
\label{sec:discussion}
From the results in table \ref{tab:results}, we see a strong benefit to inputting climate variables in regions 0, 1, and 2. We see roughly ~4\%, ~5\%, and ~11\% improvement in accuracy comparing the baseline Transformer to the best accuracy achieved by ARCDoG in regions 0, 1, and 2, respectively. Region 3 is an outlier, and using climate variables seems to be detrimental to performance. Moreover, other than for region 3, a slightly positive correlation weight appears to help the model, whereas for region 3, de-correlating the climate variables from the input seems beneficial (where not using climate input appears best). 

In \cref{fig:knn}, we see how correlating the climate variables provides the transformer a leg up over simply inputting satellite data. The nearest features of the transformer in figures \ref{fig:knn0}, \ref{fig:knn1}, \ref{fig:knn2}, \ref{fig:knn3} are much noisier and less smooth than those of the transformer with climate variables in \ref{fig:knn0clim}, \ref{fig:knn1clim}, \ref{fig:knn2clim}, \ref{fig:knn3clim}. In both rows, we note that the model learns to associate points from the source regions to those in the target regions primarily based on longitude. This roughly mirrors what we see in figure \ref{fig:climate_knn}, where the domain variables are themselves associated mostly on longitude. We note that region 3 \ref{fig:knn3clim} seems to match most strongly to points very close to its border, with relatively fewer matches from different parts of the US. This hints at why its performance is an outlier; it's geographical and climactic properties differ from those other regions, and thus domain variables (i.e. climate) appear to be insufficient to generalise. Moreover, the climate transformer in the bottom row associates points from the Pacific Northwest with those in the South East in \ref{fig:knn1clim} and \ref{fig:knn2clim}. This analysis suggests the presence of similar geographic and climactic regions in disparate parts of the US that might be amenable to growing similar types of crops.


In table \ref{tab:ablate}, we see that the temperature variables are more important to regions 0 and 2, while the precipitation variables are more important to regions 1 and 3. These results seem to follow lines of latitude, where regions 0 and 2 form the Southern half, and regions 1 and 3 the Northern half of the US. Even then, it is somewhat surprising that the model does better with only one set of variables (temperature only in regions 0 and 2, and precipitation only in region 3) than with both, demonstrating that the model is suboptimally exploiting correlations in the spectral channels of the input. 

\section{Conclusion}
\label{sec:conclusion}
In this work we demonstrate the possibility of continuous-variable domain generalisation in machine learning, applied to satellite imagery. The incorporation of a regression term to correlate domain variables with model outputs yields a simple method of domain generalisation. We demonstrate large-scale generalisation by breaking the traditional pipeline of treating satellite imagery as ``images" and instead as a grid of continually distributed feature-rich data points. Further research can investigate learning the strength of correlation dynamically, as it may be optimal to correlate some portions of the input while de-correlating others. Moreover, recent work in self-supervised methods for satellite imagery \cite{gassl} is relevant to the question of generalising to out-of-domain tasks and distributions, and may work in tandem with domain generalisation.

In terms of applications to satellite imagery, further investigation into higher resolution imagery could improve model performance on the task of crop classification. Moreover, work needs to be done to investigate an optimal architecture for large-scale satellite imagery. Even as the Transformer model demonstrates strong performance, investigating how to leverage spatial, temporal, and spectral variance in the input coherently is an open question. 

ARCDoG provides a glimpse into devising unique machine learning solutions for remote sensing. Reframing the problem of crop classification into one of domain generalisation, we demonstrate that much can be learned about both the machine learning tools and the task to which they are applied. We hope this work spurs further research into nuanced applications of computer vision for satellite imagery.

{
\small
\bibliographystyle{ieee_fullname}
\bibliography{egbib}

\begin{thebibliography}{10}\itemsep=-1pt

\bibitem{earth_engine_datasets}
Earth engine dataset catalog.
\newblock https://developers.google.com/earth-engine/datasets/.
\newblock Accessed: Nov. 10 2021.

\bibitem{landsat}
Landsat missions: Landsat 8.
\newblock https://www.usgs.gov/core-science-systems/nli/landsat/landsat-8.
\newblock Accessed: Nov. 10 2021.

\bibitem{modis}
Moderate resolution imaging spectroradiometer.
\newblock https://modis.gsfc.nasa.gov/.
\newblock Accessed: Nov. 10 2021.

\bibitem{naip}
Naip imagery.
\newblock
  https://www.fsa.usda.gov/programs-and-services/aerial-photography/imagery-programs/naip-imagery/.
\newblock Accessed: Nov. 10 2021.

\bibitem{sentinel}
Sentinel missions: Sentinel-2.
\newblock https://sentinel.esa.int/web/sentinel/missions/sentinel-2.
\newblock Accessed: Nov. 10 2021.

\bibitem{hardiness_zones}
U.s. forest service.

\bibitem{gassl}
Kumar Ayush, Burak Uzkent, Chenlin Meng, Kumar Tanmay, Marshall Burke, David~B.
  Lobell, and Stefano Ermon.
\newblock Geography-aware self-supervised learning.
\newblock {\em CoRR}, abs/2011.09980, 2020.

\bibitem{metareg}
Yogesh Balaji, Swami Sankaranarayanan, and Rama Chellappa.
\newblock Metareg: Towards domain generalization using meta-regularization.
\newblock {\em Advances in Neural Information Processing Systems},
  31:998--1008, 2018.

\bibitem{usda_cdl}
Claire Boryan, Zhengwei Yang, Rick Mueller, and Mike Craig.
\newblock Monitoring us agriculture: the us department of agriculture, national
  agricultural statistics service, cropland data layer program.
\newblock {\em Geocarto International}, 26(5):341--358, 2011.

\bibitem{jigen}
Fabio~M Carlucci, Antonio D'Innocente, Silvia Bucci, Barbara Caputo, and
  Tatiana Tommasi.
\newblock Domain generalization by solving jigsaw puzzles.
\newblock In {\em Proceedings of the IEEE/CVF Conference on Computer Vision and
  Pattern Recognition}, pages 2229--2238, 2019.

\bibitem{gan_survey}
Antonia Creswell, Tom White, Vincent Dumoulin, Kai Arulkumaran, Biswa Sengupta,
  and Anil~A Bharath.
\newblock Generative adversarial networks: An overview.
\newblock {\em IEEE Signal Processing Magazine}, 35(1):53--65, 2018.

\bibitem{deepglobe}
Ilke Demir, Krzysztof Koperski, David Lindenbaum, Guan Pang, Jing Huang, Saikat
  Basu, Forest Hughes, Devis Tuia, and Ramesh Raskar.
\newblock Deepglobe 2018: A challenge to parse the earth through satellite
  images.
\newblock In {\em Proceedings of the IEEE Conference on Computer Vision and
  Pattern Recognition Workshops}, pages 172--181, 2018.

\bibitem{worldclim}
Stephen~E Fick and Robert~J Hijmans.
\newblock Worldclim 2: new 1-km spatial resolution climate surfaces for global
  land areas.
\newblock {\em International journal of climatology}, 37(12):4302--4315, 2017.

\bibitem{canada_crops}
T Fisette, P Rollin, Z Aly, L Campbell, B Daneshfar, P Filyer, A Smith, A
  Davidson, J Shang, and I Jarvis.
\newblock Aafc annual crop inventory.
\newblock In {\em 2013 Second International Conference on Agro-Geoinformatics
  (Agro-Geoinformatics)}, pages 270--274. IEEE, 2013.

\bibitem{dann}
Yaroslav Ganin, Evgeniya Ustinova, Hana Ajakan, Pascal Germain, Hugo
  Larochelle, Fran{\c{c}}ois Laviolette, Mario Marchand, and Victor Lempitsky.
\newblock Domain adversarial training of neural networks.
\newblock {\em The journal of machine learning research}, 17(1):2096--2030,
  2016.

\bibitem{pseudoinverse}
TNE Greville.
\newblock Some applications of the pseudoinverse of a matrix.
\newblock {\em SIAM review}, 2(1):15--22, 1960.

\bibitem{hu2003climate}
Qi Hu and Gregory Buyanovsky.
\newblock Climate effects on corn yield in missouri.
\newblock {\em Journal of Applied Meteorology and Climatology},
  42(11):1626--1635, 2003.

\bibitem{almanac2020}
Sarah Janssen.
\newblock {\em The World Almanac and book of facts 2020}.
\newblock Simon and Schuster, 2019.

\bibitem{tile2vec}
Neal Jean, Sherrie Wang, Anshul Samar, George Azzari, David Lobell, and Stefano
  Ermon.
\newblock Tile2vec: Unsupervised representation learning for spatially
  distributed data.
\newblock In {\em Proceedings of the AAAI Conference on Artificial
  Intelligence}, volume~33, pages 3967--3974, 2019.

\bibitem{joachimstransductive}
Thorsten Joachims.
\newblock Transductive learning via spectral graph partitioning.
\newblock In {\em Proceedings of the 20th International Conference on Machine
  Learning (ICML-03)}, pages 290--297, 2003.

\bibitem{cifar}
Alex Krizhevsky et~al.
\newblock Learning multiple layers of features from tiny images.
\newblock Technical report, Citeseer, 2009.

\bibitem{pseudolabeling}
Dong-Hyun Lee.
\newblock Pseudo-label: The simple and efficient semi-supervised learning
  method for deep neural networks.

\bibitem{epidg}
Da Li, Jianshu Zhang, Yongxin Yang, Cong Liu, Yi-Zhe Song, and Timothy~M
  Hospedales.
\newblock Episodic training for domain generalization.
\newblock In {\em Proceedings of the IEEE/CVF International Conference on
  Computer Vision}, pages 1446--1455, 2019.

\bibitem{imagenet}
Olga Russakovsky, Jia Deng, Hao Su, Jonathan Krause, Sanjeev Satheesh, Sean Ma,
  Zhiheng Huang, Andrej Karpathy, Aditya Khosla, Michael Bernstein, et~al.
\newblock Imagenet large scale visual recognition challenge.
\newblock {\em International journal of computer vision}, 115(3):211--252,
  2015.

\bibitem{cropTransformer}
Marc Ru{\ss}wurm and Marco K{\"{o}}rner.
\newblock Self-attention for raw optical satellite time series classification.
\newblock {\em CoRR}, abs/1910.10536, 2019.

\bibitem{bigearthnet}
Gencer Sumbul, Marcela Charfuelan, Beg{\"u}m Demir, and Volker Markl.
\newblock Bigearthnet: A large-scale benchmark archive for remote sensing image
  understanding.

\bibitem{testtimetraining}
Yu Sun, Xiaolong Wang, Liu Zhuang, John Miller, Moritz Hardt, and Alexei~A.
  Efros.
\newblock Test-time training with self-supervision for generalization under
  distribution shifts.
\newblock In {\em ICML}, 2020.

\bibitem{transformers}
Ashish Vaswani, Noam Shazeer, Niki Parmar, Jakob Uszkoreit, Llion Jones,
  Aidan~N Gomez, {\L}ukasz Kaiser, and Illia Polosukhin.
\newblock Attention is all you need.
\newblock In {\em Advances in neural information processing systems}, pages
  5998--6008, 2017.

\bibitem{tempImportant}
Lucija Viskovic, Ivana~Nizetic Kosovic, and Toni Mastelic.
\newblock Crop classification using multi-spectral and multitemporal satellite
  imagery with machine learning.
\newblock In {\em 2019 International Conference on Software, Telecommunications
  and Computer Networks (SoftCOM)}, pages 1--5, 2019.

\bibitem{frn}
Davis Wertheimer, Luming Tang, and Bharath Hariharan.
\newblock Few-shot classification with feature map reconstruction networks.
\newblock In {\em Proceedings of the IEEE/CVF Conference on Computer Vision and
  Pattern Recognition}, pages 8012--8021, 2021.

\bibitem{merced}
Yi Yang and Shawn Newsam.
\newblock Bag-of-visual-words and spatial extensions for land-use
  classification.
\newblock In {\em Proceedings of the 18th SIGSPATIAL international conference
  on advances in geographic information systems}, pages 270--279. ACM, 2010.

\bibitem{mcr2}
Yaodong Yu, Kwan Ho~Ryan Chan, Chong You, Chaobing Song, and Yi Ma.
\newblock Learning diverse and discriminative representations via the principle
  of maximal coding rate reduction, 2020.

\bibitem{s4l}
Xiaohua Zhai, Avital Oliver, Alexander Kolesnikov, and Lucas Beyer.
\newblock S4l: Self-supervised semi-supervised learning.
\newblock In {\em Proceedings of the IEEE international conference on computer
  vision}, pages 1476--1485, 2019.

\bibitem{mdan}
Han Zhao, Shanghang Zhang, Guanhang Wu, Jos{\'e}~MF Moura, Joao~P Costeira, and
  Geoffrey~J Gordon.
\newblock Adversarial multiple source domain adaptation.
\newblock {\em Advances in neural information processing systems},
  31:8559--8570, 2018.

\bibitem{dg_survey}
Kaiyang Zhou, Ziwei Liu, Yu Qiao, Tao Xiang, and Chen~Change Loy.
\newblock Domain generalization: A survey.
\newblock {\em arXiv preprint arXiv:2103.02503}, 2021.

\end{thebibliography}
}

\clearpage

\appendix 
\section{Appendix}
A few additional details about training and the Transformer architecture are mentioned here. 

\subsection{Data}
As mentioned in \cref{subsec:data}, we selected data points from 25 crop classes, discarding all others. The 25 classes were: \\

\noindent \texttt{Corn, Soybeans, Rice, Alfalfa, Grapes, Almonds, Pecans, Peanuts, Walnuts, Potatoes, Oats, Cotton, Dry beans, Sugarbeets, Winter Wheat, Spring Wheat, Durum Wheat, Sorghum, Canola, Barley, Sunflower, Pop or Orn Corn, Other Hay-Non Alfalfa, Woody Wetlands, Fallow-Idle Cropland} \\

These classes consist of some of most prevalent \textit{crop} classes, excluding land-cover classes such as shrubland, pasture, developed space, water etc which are not relevant to climate indicators. 

\subsection{Transformer Architecture}
The Transformer was implemented in PyTorch. The source code used is available \href{https://github.com/samar-khanna/cropmap/blob/04ddca8d6125cd341ff6290150998a00f0b2baf3/domain_gen/transformer.py}{here}.

We first pass the input through a layer norm, a convolutional feature extractor, and then to the Transformer layers. The convolutional feature extractor consists of a sequence of 2 blocks, where each block is a 1d convolutional layer followed by a ReLU. The feature extractor maps the 28 (or 9, in the case of the baseline) input channels to an intermediate feature dimension of 64. Another layer normalisation is applied, before passing the features to a positional encoding layer (on the temporal dimension), and then to a Transformer encoder with 2 encoder layers, each with 2 attention heads, dropout, ReLU, and a feedforward dimension of 256. Another layer normalisation is used, before the outputs for each of the 8 time points are max pooled and passed through a final 1d convolutional layer with 25 channels, for classification. The features used for \cref{fig:knn} are from the output of the max pool layer.

\end{document}